\documentclass[sigconf]{acmart}

\AtBeginDocument{%
  }

\copyrightyear{2024}
\acmYear{2024}
\setcopyright{rightsretained}
\acmConference[CIKM '24]{Proceedings of the 33rd ACM International Conference on Information and Knowledge Management}{October 21--25, 2024}{Boise, ID, USA}
\acmBooktitle{Proceedings of the 33rd ACM International Conference on Information and Knowledge Management (CIKM '24), October 21--25, 2024, Boise, ID, USA}
\acmDOI{10.1145/3627673.3679837}
\acmISBN{979-8-4007-0436-9/24/10}

\usepackage{amsmath,amsfonts}
\usepackage{algorithmic}
\usepackage{algorithm}
\usepackage{array}
\usepackage{booktabs}
\usepackage[caption=false,font=scriptsize,labelfont=sf,textfont=sf]{subfig}
\usepackage{microtype}
\usepackage{multirow}
\usepackage{makecell}
\usepackage{textcomp}
\usepackage{stfloats}
\usepackage{url}
\usepackage[normalem]{ulem}
\usepackage{verbatim}
\usepackage{graphicx}
\usepackage{color,xcolor}

\newcommand{\modelName}{GenExpan}

\begin{document}

\title{From Retrieval to Generation: Efficient and Effective\\Entity Set Expansion}

\author{Shulin Huang}
\authornote{Both authors contributed equally to this research.}
\orcid{0009-0006-2181-8073}
\affiliation{%
  \institution{Shenzhen International Graduate School, Tsinghua Universiy}
  \city{Shenzhen}
\country{China}
}
\email{sl-huang21@mails.tsinghua.edu.cn}

\author{Shirong Ma}
\authornotemark[1]
\orcid{0009-0008-1686-407X}
\affiliation{%
  \institution{Shenzhen International Graduate School, Tsinghua Universiy}
    \city{Shenzhen}
\country{China}
}
\email{masr21@mails.tsinghua.edu.cn}

\author{Yangning Li}
\orcid{0000-0002-1991-6698}
\affiliation{%
  \institution{Shenzhen International Graduate School, Tsinghua Universiy}
      \city{Shenzhen}
\country{China}
}
\email{liyn20@mails.tsinghua.edu.cn}

\author{Yinghui Li}
\orcid{0000-0001-7571-6722}
\affiliation{%
  \institution{Shenzhen International Graduate School, Tsinghua Universiy}
      \city{Shenzhen}
\country{China}
}
\email{liyinghu20@mails.tsinghua.edu.cn}

\author{Hai-Tao Zheng}
\orcid{0000-0001-5128-5649}
\affiliation{%
  \institution{Shenzhen International Graduate School, Tsinghua Universiy}
      \city{Shenzhen}
\country{China}
}
\affiliation{%
  \institution{Pengcheng Laboratory}
  \city{Shenzhen}
\country{China}
}
\authornote{The corresponding author.}
\email{zheng.haitao@sz.tsinghua.edu.cn}

\renewcommand{\shortauthors}{Shulin Huang et al.}

\begin{abstract}
Entity Set Expansion (ESE) is a critical task aiming at expanding entities of the target semantic class described by seed entities.
Most existing ESE methods are retrieval-based frameworks that need to extract contextual features of entities and calculate the similarity between seed entities and candidate entities.
To achieve the two purposes, they iteratively traverse the corpus and the entity vocabulary, resulting in poor efficiency and scalability.
Experimental results indicate that the time consumed by the retrieval-based ESE methods increases linearly with entity vocabulary and corpus size.
In this paper, we firstly propose \textbf{Gen}erative Entity Set \textbf{Expan}sion (\textbf{GenExpan}) framework, 
which utilizes a generative pre-trained auto-regressive language model to accomplish ESE task.
Specifically, a prefix tree is employed to guarantee the validity of entity generation, and automatically generated class names are adopted to guide the model to generate target entities.
Moreover, we propose Knowledge Calibration and Generative Ranking to further bridge the gap between generic knowledge of the language model and the goal of ESE task.
For efficiency, expansion time consumed by \modelName{} is independent of entity vocabulary and corpus size, and \modelName{} achieves an average 600\% speedup compared to strong baselines.
For expansion effectiveness, our framework outperforms previous state-of-the-art ESE methods.
\end{abstract}





\maketitle

\section{Introduction}
Entity Set Expansion (ESE) is a critical task that aims to expand a small set of seed entities (e.g., {``United States'', ``China'', ``Britain''}) with new entities (e.g., {``Canada'', ``Japan'', ``Australia''}) that belongs to the same target semantic class (i.e., Country)~\cite{shi2014probabilistic}.
Thanks to the ability to mine the semantics, ESE task benefits various downstream NLP and IR applications~\cite{zhao2018entity}, such as Knowledge Graph~\cite{shi2021entity, DBLP:journals/corr/abs-2211-04215, DBLP:journals/corr/abs-2302-08774}, Question Answering~\cite{wang2008automatic}, and Semantic Search~\cite{xiong2017explicit,chen2016long}.

Recently, mainstream approaches for ESE task have predominantly revolved around corpus-dependent, iterative, retrieval-based methods~\cite{zhang2020empower, li2022contrastive, yu2019corpus, li2023automatic}. These methods assume that similar entities have similar semantic representations in contexts~\cite{zhang2018entity,barrena-etal-2016-alleviating}.
In ESE task, ``corpus'' specifically denotes a collection of sentences where candidate entities occur.
Retrieval-based ESE methods extract and model contextual features of candidate entities from a large-scale corpus to represent entities. 
These methods then iteratively calculate the similarity between seed and candidate entities, selecting a few entities with the highest similarity as the expansion results. 


\begin{figure}
    \centering
    \includegraphics[width=0.9\columnwidth]{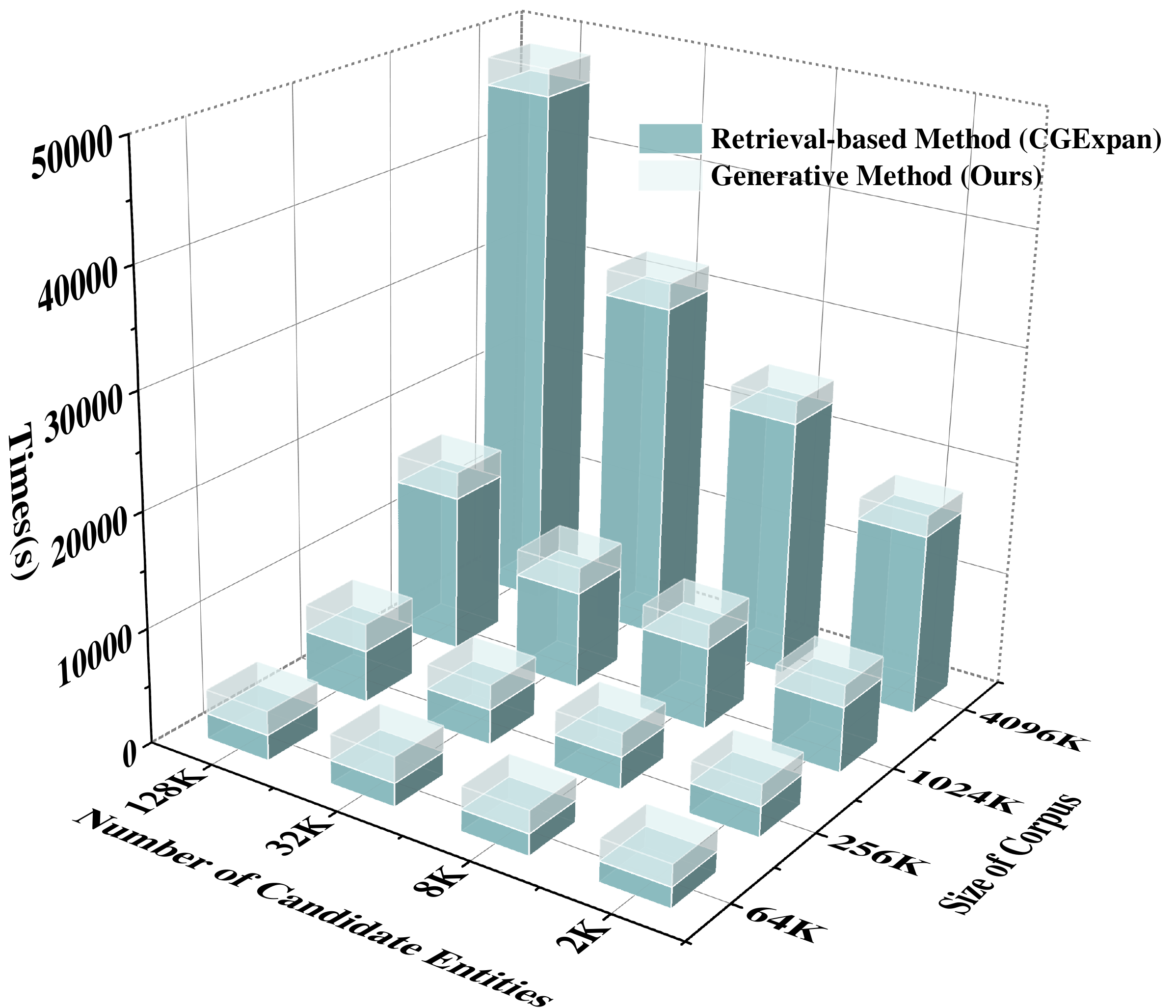}
    \caption{
    Total elapsed time for retrieval-based and generative methods. We select strong baseline CGExpan to represent retrieval-based methods. All experiments are run on one Nvidia RTX 3090 GPU.
    The two horizontal axes represent the number of candidate entities and corpus size, i.e., the number of sentences in the corpus.
    The vertical axis represents total elapsed time of entire expansion process. We execute 40 queries, with each expanding a minimum of 50 entities.
    }

    \label{fig:intro}
\end{figure}

Unfortunately, 
such retrieval-based approaches lack time efficiency and scalability.
On one hand, they traverse the large corpus to extract the contextual features of entities.
Thus the time consumption achieves $\mathcal{O}(N)$ where $N$ is the number of sentences in the corpus.
On the other hand, 
they rank candidates by calculating similarity between seed and candidate entities during each iteration.
Thus the time consumption of each iteration achieves at least $\mathcal{O}(|S|*|V|)$, where $S$ and $V$ are seed entity set and candidate entity vocabulary, respectively.
As shown in Figure~\ref{fig:intro}, the time required to expand at least 50 entities for a retrieval-based method increases sharply with the number of candidates and corpus size. This makes existing retrieval-based methods impractical for large corpus, e.g., the complete Wikipedia contains more than 6M pages.

To address the above problems, we propose a novel corpus-independent generative ESE framework for the first time, namely \textbf{Gen}erative Entity Set \textbf{Expan}sion (\textbf{\modelName{}}). Our approach utilizes an auto-regressive language model pre-trained on large-scale text~\cite{radford2019language, brown2020language, zhang2022opt} to accomplish ESE task without fine-tuning the model or extracting features from the corpus.
However, there is a gap between the generic knowledge within the pre-trained language model and the target of ESE task. 
The language model is pre-trained on language modeling task, 
while ESE task requires the model to generate entities belonging to the same semantic class as given seed entities.
Therefore, we should carefully devise the generation process to guide the model to generate target entities and rank them according to their similarity to seed entities.

Specifically, \modelName{} consists of 4 parts:
(1) \emph{Class Name Generation.} We adopt In-context Learning to guide the model in generating the semantic class name of seed entities. 
The generated class name is incorporated into the prompt used for entity generation. 
(2) \emph{Prefix-constrained Entity Generation.} 
We construct a prompt based on given seed entities and generated class names, feed it to the language model, and employ a prefix tree constructed by the candidate entity vocabulary to constrain generated tokens.
(3) \emph{Knowledge Calibration.} We obtain the ``prior probability'' of the language model by inputting a meaningless prompt to the model, and employ it to adjust the output probability distribution during the generation process, which reduces over-preference of the model for common entities.
(4) \emph{Generative Ranking.} We design templates that generate seed entities and the class name from target entities in reverse, and calculate log probability of the language model generating the templates.
The log probability is used to measure semantic similarity between generated entities and seed entities, helping to further rank the generated entities.

Throughout the expansion process, we employ the same pre-trained language model to generate different texts, with the generation time proportional to the text length.
Specifically, in \modelName{}, it is related to the length of pre-defined prompts, the length of candidate entities, and the length of class names.
Hence, the time consumed by our generative approach is approximately $\mathcal{O}(L_p+L_c+L_e)$ where $L_p, L_c, L_e$ are the length of prompts, class names and entities, respectively.
Results in Figure~\ref{fig:intro} demonstrate that unlike previous corpus-dependent retrieval-based methods, the elapsed time of our approach is independent of entity vocabulary and corpus size.

Extensive experiments on four publicly available datasets demonstrate the effectiveness and efficiency of our approach: 
(1) \modelName{} outperforms state-of-the-art methods in all evaluation metrics for different datasets.
(2) \modelName{} expands entities faster than various previous iterative retrieval-based methods on different datasets, achieving an average speedup of 600\% compared to strong baselines.

\section{Related Work}

\subsection{Entity Set Expansion Methods}
Due to its wide-ranging applications\cite{pantel2009web,shen2020taxoexpan,yu2020learning,li2024mesed,li2024ultrawiki}, Entity Set Expansion (ESE) has attracted substantial attention.
Its semantic exploration and knowledge discovery capabilities are valuable for tasks like
Web Search~\cite{xiong2017explicit,chen2016long,koumenides2014ranking}, Taxonomy~\cite{velardi-etal-2013-ontolearn,treeratpituk2006automatically,yin2010building}, Recommendation~\cite{zhao2017entity,blanco2013entity,yu2014personalized} and Question Answering~\cite{wang-etal-2008-automatic,wang2008automatic,dubey2018earl}.

Early research in ESE task primarily focuses on web information~\cite{wang2007language,tong2008system}, relying on search engines and web pages for entity expansion, which led to low efficiency. 
Retrieval-based methods are crucial in entity-related tasks, such as Entity Set Expansion, Entity set naming~\cite{huang2021autoname} and Entity Linking~\cite{sevgili2022neural}.
Retrieval-based methods have recently become the mainstream of ESE~\cite{yu2019corpus,kushilevitz2020two,zhang2020empower}. 
Early retrieval-based frameworks \cite{kushilevitz2020two,mamou2018term,yu2019corpus} rank entities according to their distributional similarity in the corpus and complete the expansion only in one sort. Later retrieval-based frameworks\cite{huang2020guiding,rong2016egoset,shen2017setexpan} bootstrap seed entities set by iteratively ranking entities according to context features.
Particularly, CGExpan\cite{zhang2020empower} leverages Hearst patterns to construct probing queries for entity presentation and utilizes semantic information from corpus. SynSetExpan\cite{shen2020synsetexpan} adds the synonym information~\cite{DBLP:journals/corr/abs-2211-10997} to give more supervision signals in expansion. 
ProbExpan\cite{li2022contrastive} devises an entity-level masked language model with contrastive learning~\cite{DBLP:conf/acl/LiZLLLSWLCZ22} to refine the representation of entities, achieving current state-of-the-art performance.

However, all recent methods rely on corpus and their time consumption grows linearly with entity vocabulary and corpus size.
In contrast, we propose a corpus-independent generative framework that fully exploits pre-trained model's intrinsic knowledge without using the corpus, ensuring both effective and efficient expansion.

\subsection{Entity Set Expansion Resources}
In previous works, there are some traditional data resources for Entity Set Expansion. 
\textbf{APR}~\cite{shen2017setexpan} provides a large-scale corpus collected from news articles published by Associated Press and Reuters in 2015. \textbf{Wiki}~\cite{shen2017setexpan} provides the corpus, which is a subset of English Wikipedia articles. These two datasets contain 3 and 8 semantic categories, respectively.
\textbf{CoNLL}~\cite{zupon2019lightly} is constructed from the CoNLL 2003 shared task dataset~\cite{sang1837introduction} and contains 4 categories (5,522 entities). 
\textbf{OntoNotes}~\cite{zupon2019lightly} is constructed from OntoNotes datasets~\cite{pradhan2013towards} and contains 11 categories (19,984 entities).
For CoNLL and OntoNotes, the n-grams of up to 4 tokens on either side of an entity are used as patterns, and some patterns are filtered out to construct datasets for ESE task~\cite{zupon2019lightly}.
CoNLL and OntoNotes datasets are constructed from NER datasets which contain domains different from APR and Wiki datasets, posing a challenge to the generalizability of ESE methods.
In this paper, we use all these four datasets to verify the effectiveness and the efficiency of our framework \modelName{}, which also shows the generalizability of our framework.

\begin{figure*}[ht]
\centering
\includegraphics[width=0.8\textwidth]{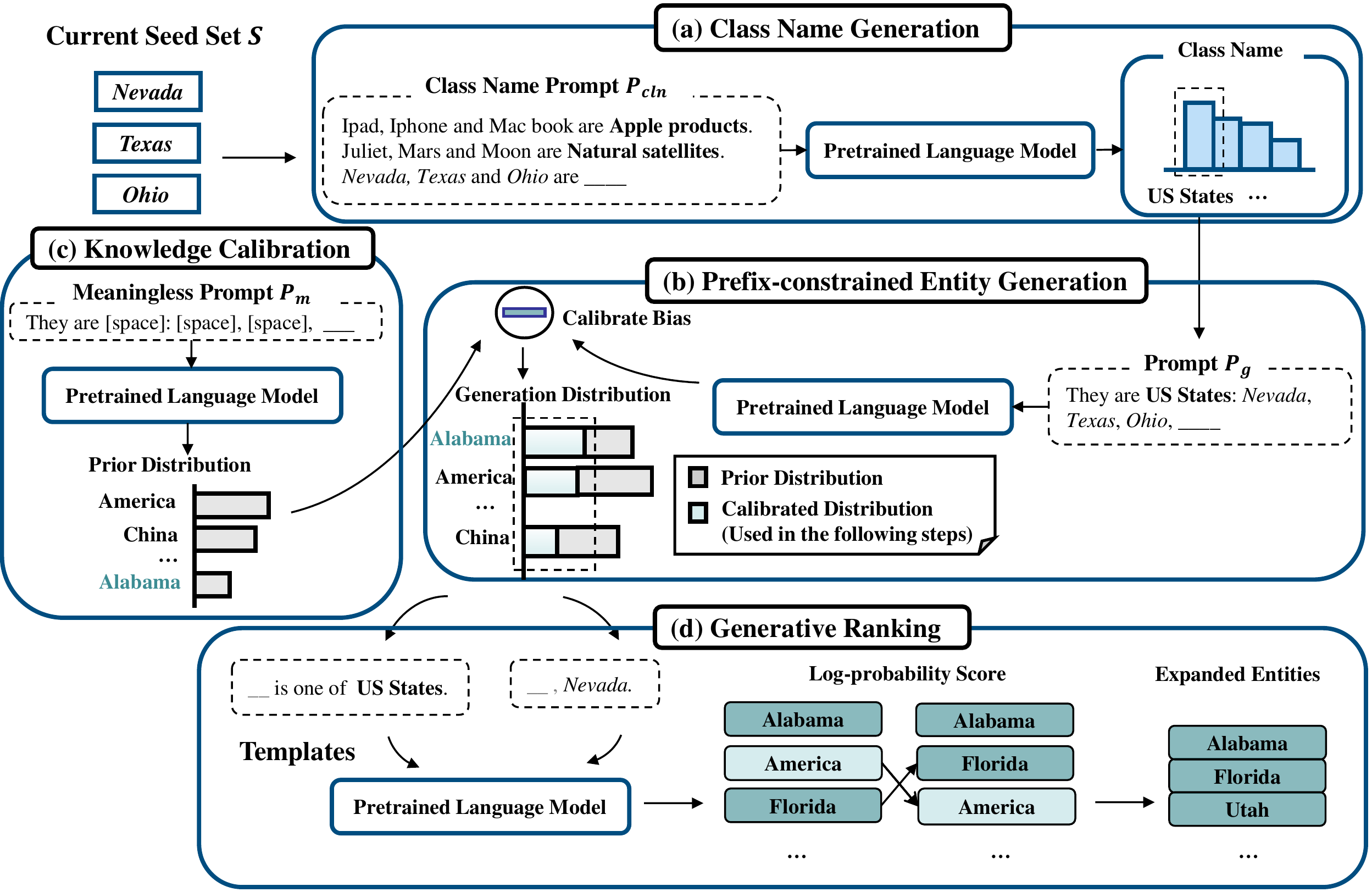}
\caption{Overview of \modelName{} framework. We employ the same pre-trained language model during whole expansion process.}
\label{Method_Figure}
\end{figure*}

\section{Methodology}

\subsection{Problem Formulation}
Entity Set Expansion (ESE) aims to identify entities that belong to the same semantic class as several entities provided by the user.
Given a small seed entity set $S=\{s_1,s_2,...,s_n\}$ consisting of several similar entities and an entity vocabulary $V$ containing all candidate entities, it is expected to obtain an entity list $E\subset V$. Each entity in the list should belong to the same semantic class as all entities in $S$, and the list is sorted by semantic similarity from highest to lowest.

In addition, a large-scale corpus $D$ is also given in previous work, which contains the context of each entity in the entity vocabulary $V$. While previous retrieval-based methods model the contextual representations of entities from the corpus $D$, our proposed generative framework would not require the corpus.

\subsection {Overview of Methodology}\label{sec:overall}
The overview of \modelName{} is shown in Figure~\ref{Method_Figure}.
Our approach employs an auto-regressive model to address ESE task, comprising four components: Class Name Generation, Prefix-constrained Entity Generation, Knowledge Calibration, and Generative Ranking.

In Class Name Generation, we employ In-context Learning to guide language model in generating the class name (Section~\ref{sec:cln}).
In Prefix-constrained Entity Generation, 
we construct a prompt with seed entities and generated class name, feeding it into auto-regressive model to generate target entities. 
To ensure that the model generates only valid entities within the entity vocabulary, we build a prefix tree based on the entity vocabulary and employ prefix-constrained Beam Search for decoding (Section~\ref{sec:gen}).

Moreover, we propose two additional steps, 
to further bridge the gap between the generic knowledge implicit in the language model and the target of ESE task.
In Knowledge Calibration, the output probability distribution during model generation is adjusted so as to reduce the model's over-preference for common entities (Section~\ref{sec:cal}).
In Generative Ranking, we devise templates that generate seed entities and the class name from target entities, allowing us to measure the semantic similarity between target entities and seed entities (Section~\ref{sec:gr}).
The same language model is used for the entire expansion process, without any fine-tuning.
\begin{figure*}[ht]
\centering
\subfloat[Outside entity generation: 
The current prompt ends with a comma, indicating outside the entity generation step. Therefore, we should start to generate an entity from entities prefix tree $T$, e.g., both ``Florida'' and ``China'' are valid beginning tokens in $T$.]
{ \label{fig:tr1} 
\includegraphics[width=0.65\columnwidth]{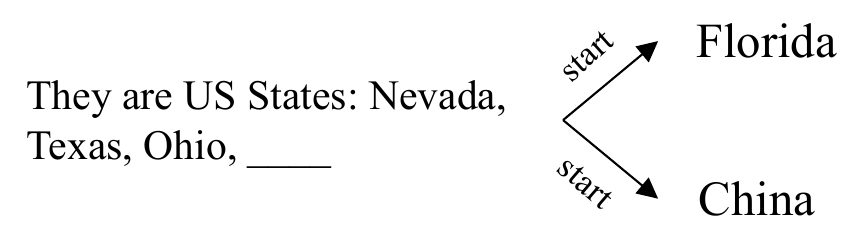} 
}
\qquad
\subfloat[Inside entity generation: After some tokens are generated (e.g., \text{Florida}), there are two available choices. We can either continue to generate the remaining tokens of the entity if an entity is not completely generated (e.g., ``Florida State''), or terminate the generation process using commas if the generated prefix already constitutes a valid entity (e.g., ``Florida'').] 
{ \label{fig:tr2} 
\includegraphics[width=0.65\columnwidth]{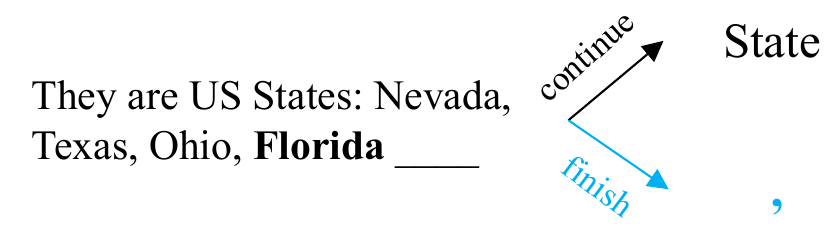} 
} 
\caption{Example of constrained entity decoding using ``They are US States: Nevada, Texas, Ohio,'' as input. There are 2 cases: when we are outside entity generation (a), inside entity generation (b). The model is supposed to generate valid entities from entities prefix tree $T$ (e.g., ``China'', ``Florida'', ``Florida State'')  based on the input.}
\label{fig:tr}
\end{figure*}

\subsection{Class Name Generation}\label{sec:cln}
With sufficient pre-training on large-scale texts, the auto-regressive language model is equipped with powerful text generation capabilities~\cite{DBLP:journals/csur/DongLGCLSY23, DBLP:conf/icassp/SunCZLCZ22,  DBLP:journals/patterns/LiuLTLZ22}. The language model tends to generate fluent and complete sentences during the generation process~\cite{DBLP:journals/corr/abs-2210-16557}. However, the ESE task expects the model to generate entities that belong to the same semantic class as given seed entities. It means a significant gap exists between the ESE task and the text generation task.


In order to enhance the effectiveness of the model for ESE task, we apply In-context Learning, which involves a class name prompt to guide the model in generating the semantic class name of seed entities before generating target entities.
Specifically, we design a class name prompt $y_{c}=<{s}_{1},~{s}_{2},...~\text{and}~{s}_{n}~\text{are}>$ and make the model output resemble a natural language sentence: ${s}_{1},~{s}_{2},...~\text{and}~{s}_{n}~\text{are}~ c$, where $c$ is the generated class name. To guide the model to better understand the task goal of class name generation, we preface class name prompt $y_c$ with a few examples of manually set generated class names. 
To avoid data leakage, the provided examples do not include any semantic classes from all datasets used for experiments. For instance, the constructed class name prompt for seed entities (\emph{``Nevada''}, \emph{``Texas''}, \emph{``Ohio''}) is ``IPad, ~Iphone and MacBook Pro are Apple products.~Juliet,~Mars and Moon are Natural satellites.~Nevada,~Texas and Ohio are''. Guided by the given examples, the model is capable of summarising the common category to which seed entities belong (e.g., U.S. states).
Once the class name is obtained, it is integrated into the prompt for the subsequent step of entity generation, which imposes more stringent constraints.

\subsection{Prefix-constrained Entity Generation}\label{sec:gen}
In our approach, ESE is treated as an auto-regressive text generation task. We utilize the given seed entity set $S=\{s_1,s_2,...,s_n\}$ and the class name $c$ generated in the previous step to construct a prompt. This prompt is used as a prefix input to a pre-trained auto-regressive language model, which enables the model to generate entities belonging to the same semantic class as the entities in $S$.
Specifically, we adopt a straightforward approach for constructing the prompt, which involves concatenating the entities in $S$ (e.g., \emph{``Nevada''}, \emph{``Texas''}, \emph{``Ohio''}) using commas, and inserting the class name information $c$ to the beginning of the prompt.
This prompt can be represented as follows: $y_{g} =<\text{They are }c \text{: } s_1, s_2,..., s_n>$, e.g., ``They are \emph{US states}: \emph{Nevada}, \emph{Texas}, \emph{Ohio},''.

However, utilizing the language model for generation without modification results in generating irrelevant content instead of expected entities. To force the model to generate valid entities, i.e., entities in vocabulary $V$, we construct a prefix tree $T$. Each leaf node of this tree corresponds to an entity in $V$, and a path from the root node to a leaf node represents all tokens that constitute a complete entity. For each node $t\in T$, its children represent succeeding tokens allowed to be generated after the prefix corresponding to the node $t$.
We exploit prefix-constrained Beam Search~\cite{sutskever2014sequence} as the decoding strategy. At each step, prefix tree $T$ is employed to restrict the next decoded token by setting the decoding probability of invalid tokens to 0, thus forcing the model to generate entities in $V$.
Figure~\ref{fig:tr} presents an example of prefix-constrained entity generation.

The decoding process of one beam in Beam Search is terminated after a complete entity is generated. Those decoded entities $E=\{e_1,e_2,...,e_K\}$ are sorted in descending order of their scores in terms of decoding probability output by the language model:
\begin{equation}
    \text{score}(e_j|S,c)=\log{p_{\theta}(y|y_g)}=\prod^N_{i=\left|y_g\right|+1}p_{\theta}(y_i|y_{<i}),
\end{equation}
where $y$ denotes a complete sequence consisting of $N$ tokens generated by the language model, while $\theta$ represents parameters of the model. Notably, the generated sequence $y$ encompasses the prompt $y_g$ fed into the model, and a decoded entity $e_j$.

Parallel decoding of Beam Search ensures high computational efficiency of \modelName{}.
Since the auto-regressive approach is utilized to generate target entities, the time consumed for entity generation depends on the size of the beams $b$, the length of entities (e.g., the entities of Wiki have 4.6 BPE tokens on average) and input prefix, which is independent of the size of entity vocabulary $V$.

\subsection{Knowledge Calibration}\label{sec:cal}

Pre-trained auto-regressive language models such as GPT-2~\cite{radford2019language} contain extensive general world knowledge. However, our observation reveals that, when implementing the method mentioned in Section~\ref{sec:gen}, model tends to generate more common entities rather than the entities that are more semantically similar to seed entities.

Inspired by calibration in classification task~\cite{zhao2021calibrate}, we design Knowledge Calibration to adjust the output distribution of the model to reduce generation bias.
Concretely, we replace the class name and specific entities in the prompt mentioned in Section~\ref{sec:gen} with spaces, transforming the prompt into the following form (also called a meaningless prompt) $y_{m}=$ <They are [blank]: [blank], [blank], [blank],>. [blank] serves as a meaningless token that can be a blank character such as a space. We use such a meaningless prompt for the pre-trained language model and obtain the probability distribution of the next token output by the model.

Since the meaningless prompt does not contain any semantic information about specific entities or categories,
in an ideal scenario, the output probability of the model should be similar for tokens corresponding to different entities.
In practice, the pre-trained language model tends to assign significantly higher probability to tokens associated with common entities compared to uncommon tokens.
Here, we refer to this probability distribution as ``prior probability'' $\hat{p}$ of the model.
During the decoding process, we adopt the prior probability $\hat{p}$ to modify the probability distribution as follows to reduce the effect of commonness of entities on model generation:
\begin{equation}
    p'(y)=p(y)\cdot\hat{p}^{-\mu},
\end{equation}
where $p$ denotes the probability distribution of the next token from the pre-trained language model and $p'$ denotes the score after calibration.
$\mu\in[0,1]$ is a hyper-parameter.

Knowledge Calibration enables the model to avoid over-preference for commonplace entities and consequently bridges the gap between the pre-trained language model's general knowledge and the specific task-related knowledge required for the ESE task.

\subsection{Generative Ranking}\label{sec:gr}

Multiple reasonable entities can be generated in parallel using the generative approach mentioned in Section~\ref{sec:gen}, but the number of entities generated by one round of Beam Search is insufficient.
Following previous ESE work~\cite{zhang2020empower,shen2020synsetexpan,li2022contrastive}, we adopt an iterative approach to generate more entities: at the beginning of each iteration, we randomly select $m$ permutations composed of all seed entities, and at the end of an iteration, the top-$k$ newly generated entities are added to the seed entity set. The expansion process executes iteratively until the seed entity set contains sufficient entities.


Here, we calculate the ranking of an entity by combining the number of occurrences of an entity and its initial ranking during the Beam Search decoding. Specifically, assuming that the generation probability of entity $e$ is ranked at $r_i(e)\in\{1,2,...,K\}$ in the $i$-th iteration, the combined score $M_1(e)$ of the entity in $N$ iterations is represented as the sum of Mean Reciprocal Rank (MRR):
\begin{equation}
    M_1(e)=\sum^N_{i=1}\dfrac{1}{r_i(e)}.
\end{equation}

However, this entity ranking method still has considerable shortcomings. The ranking of an entity by this method is greatly influenced by the number of entity occurrences, but it cannot faithfully reveal the semantic similarity between generated entities and seed entities.
Therefore, we propose a novel generative ranking approach that measures the semantic similarity between entities to be ranked and seed entities. Specifically, we compare the probability of generating seed entities and the class name by inputting different target entities.
We devise two different templates: $<{e}_{\text{g}},~{e}_{\text{i}}>$ and $<{e}_{\text{g}} \text{ is one of } c>$, where ${e}_{\text{g}}$ is generated entities and ${e}_{\text{i}}$ is one of the seed entities.
Entities to be ranked are filled into the templates one by one.
Given an entity to be ranked, log probability for generating a seed entity or the class name is represented as follows:
\begin{equation}
    L(y|e)=-\dfrac{1}{N}\sum^N_{i=M+1}\log{p(y_i|y_{<i})},
\end{equation}
where $N$ and $M$ are lengths of the template and the entity, respectively. 
Calculating the log probability of these two templates is the answer to the ranking from the model's perspective. 
We sort the log probability computed by the two templates for different entities separately to obtain MRR scores $M_2(e)$ and $M_3(e)$ for entity $e$.

Finally, the three MRR scores $M_1(e), M_2(e), M_3(e)$ given by the above two ranking approaches are weighted summed to obtain the final score $C(e)$ for entity ranking:
\begin{equation}
    C(e)=(1-\lambda) M_1(e)+\lambda(M_2(e) + M_3(e)),
\end{equation}
where $\lambda$ is a hyper-parameter representing the ranking weight.

\section{Experiments}
\subsection{Experiment Setup}
\noindent\textbf{1. Dastasets.}
We conduct experiments on four publicly available datasets. Statistics of these datasets are summarized in Table~\ref{tab:datastat}.

\begin{table}[ht]
\small
\centering
\caption{Statistics of datasets used in our experiments.}
\scalebox{0.9}{
\begin{tabular}{l|cccc}
\toprule
 & \textbf{Wiki} & \textbf{APR} & \textbf{CoNLL} & \textbf{OntoNotes} \\ \midrule
\# Semantic Classes & 8 & 3 & 4 & 11 \\
\# Queries per Class & 5 & 5 & 1 & 1 \\
\# Seed Entities per Query & 3 & 3 & 10 & 10 \\
\# Candidate Entities & 33K & 76K & 6K & 20K \\
\# Sentences of Corpus & 973K & 1043K & 21K & 144K \\
\bottomrule
\end{tabular}
}
\label{tab:datastat}
\end{table}

\begin{table*}[ht]
    \centering
        \caption{MAP@K(K=10/20/50) of various methods on four datasets. For fair comparison, experimental results for all methods except BootstrapNet and BootstrapGAN on Wiki and APR are directly from other published papers (marked with *). The remaining results are obtained by executing the corresponding open-source codes. We leave some blanks, since the corresponding papers do not provide codes. We underline the previous state-of-the-art performance. \modelName{} outperforms the strongest baselines significantly with paired t-test considering MAP@10/20/50 ($\dag$ denotes $\text{p}<0.05$ and $\ddag$ denotes $\text{p}<0.01$).}
    \scalebox{0.82}{
    \begin{tabular}{lcccccccccccc}
    \toprule \multirow{2}{*} { \textbf{Methods} } & \multicolumn{3}{c} { \textbf{Wiki} } & \multicolumn{3}{c} { \textbf{APR} } & \multicolumn{3}{c} { \textbf{CoNLL} } & \multicolumn{3}{c} { \textbf{OntoNotes} }\\ 
    \cmidrule(r){2-4} \cmidrule(r){5-7} \cmidrule(r){8-10} \cmidrule(r){11-13}
    & K = 10 & K = 20 & K = 50 & K = 10 & K = 20 & K = 50 & K = 10 & K = 20 & K = 50 & K = 10 & K = 20 & K = 50 \\
    \midrule 
    Egoset & 90.4* & 87.7* & 74.5* & 75.8* & 71.0* & 57.0* & - & - & - & - & - & - \\
    SetExpan & 94.4* & 92.1* & 72.0* & 78.9* & 76.3* & 63.9* & 80.7 & 76.7 & 75.3 & 46.4 & 34.1 & 28.0 \\
    SetExpander & 49.9* & 43.9* & 32.1* & 28.7* & 20.8* & 12.0* & - & - & - & - & - & - \\
    CaSE & 89.7* & 80.6* & 58.8* & 61.9* & 49.4* & 33.0* & 78.1 & 75.4 & 65.5 & 52.4 & 40.9 & 31.1 \\
    BootstrapNet & 70.5 & 56.6 & 41.8 & 73.9 & 68.0 & 47.8 & 93.9 & 90.3 & 84.9 & 62.0 & 53.6 & 45.6 \\
    BootstrapGAN & 72.7 & 59.0 & 46.4 & 79.0 & 72.2 & 58.5 & \uline{97.5} & \uline{94.1} & \uline{91.3} & \uline{73.8} & \uline{64.0} & \uline{53.2} \\
    CGExpan  & 99.5* & 97.8* & 90.2* & 99.2* & 99.0* & 95.5* & 82.0 & 76.5 & 67.4 & 52.9 & 44.8 & 35.2 \\
    SynSetExpan & 99.1* & 97.8* & 90.4* & 98.5* & 99.0* & \uline{96.0}* & - & - & - & - & - & - \\
    ProbExpan  & \uline{99.5}* & \uline{98.3}* & \uline{92.9}* & \uline{100.0}* &\uline{99.6}* & 95.5* & 96.3 & 90.1 & 83.7 & 68.3 & 61.7 & 50.6 \\
    \midrule 
    \modelName{} (Ours) & \textbf{99.8}$\dag$ & \textbf{99.0}$\dag$ & \textbf{93.5}$\dag$ & \textbf{100.0} & \textbf{99.8}$\dag$ & \textbf{96.9}$\dag$ & \textbf{100.0}$\ddag$ & \textbf{95.0}$\dag$ & \textbf{93.4}$\ddag$ & \textbf{77.3}$\ddag$ & \textbf{70.1}$\ddag$ & \textbf{56.3}$\ddag$ \\
    \midrule
    \end{tabular}
    }
    \label{tab:allresulta}
\end{table*}

(1) \textbf{APR} and \textbf{Wiki}~\cite{shen2017setexpan}, which contain 3 and 8 semantic categories respectively. 
Each category consists of 5 queries, and each query has a seed set of 3 entities.
APR contains all 2015 year's news articles that are published by Associated Press and Reuters, including Countries, Parties and US States. Wiki is a subset of English Wikipedia articles, including China Provinces, Companies, Countries, Diseases, Parties, Sports Leagues, TV Channels and US States.

(2) \textbf{CoNLL} and \textbf{OntoNotes}~\cite{zupon2019lightly},  which contain 4 and 11 semantic categories respectively. Each semantic category has 10 entities that are used as the seeds, following previous work.

APR and Wiki datasets are usually used to evaluate  previous ESE methods. 
However, CoNLL and OntoNotes datasets are originally constructed from NER datasets so that they contain some general domains different from APR and Wiki datasets. General domains' nature in CoNLL and OntoNotes datasets leads to the challenge of generalizing the semantic of each category. 

To assess the performance and the generalizability of our framework \modelName{}, we use all four datasets in our experiments.


\noindent\textbf{2. Compared Methods.}
In our experiments, we select the following corpus-dependent retrieval-based methods as baselines. (1) \textbf{Egoset}~\cite{rong2016egoset}: A set expansion method that utilizes context features and word embeddings. Capturing words with similar semantics that are used to expand seed entities. (2) \textbf{SetExpan}~\cite{shen2017setexpan}: A framework based on distribution similarity and iterative context features selection. (3) \textbf{SetExpander}~\cite{mamou2018term}: A method that utilizes the classifier's predictive capabilities to determine whether the candidate entity belongs to the target semantic category. The context features are the judgments of the classifier. (4) \textbf{CaSE}~\cite{yu2019corpus}: An unsupervised corpus-based set expansion framework that leverages lexical features as well as distributed representations of entities for the set expansion task. (5) \textbf{CGExpan}~\cite{zhang2020empower}: A framework that uses BERT~\cite{devlin-etal-2019-bert} and Hearst patterns to get context features of the entities and iteratively scores the entities according to the context features. (6) \textbf{SynSetExpan}~\cite{shen2020synsetexpan}:
 A method that harnesses the power of synonym information to enhance the supervision signals throughout expansion process. (7) \textbf{ProbExpan}~\cite{li2022contrastive}: Current state-of-the-art method on Wiki/APR. It devises an entity-level masked language model with contrastive learning to refine the
representation of entities.
(8) \textbf{BootstrapNet}~\cite{yan2020global}: A method that takes the weakly-supervised learned boundaries into account, and constructs an end-to-end bootstrapping model. (9) \textbf{BootstrapGAN}\cite{yan2021progressive}: A method that employs the multi-view learning algorithm during its pre-training phase.


\noindent\textbf{3. Evaluation Metrics.}
The goal of ESE task is to expand a list of entities ranked from highest to lowest similarity to given seed entities.
Previous work~\cite{zhang2020empower, li2022contrastive} utilizes MAP@$K$, i.e., Mean Average Precision at different top-$K$ positions, as the evaluation metric.
We also employ MAP@$K$ to evaluate ESE approaches on all datasets.
MAP@$K$ is calculated as follows:
\begin{equation}
    \mbox{MAP@}K=
    \frac{1}{|Q|}\sum_{q \in Q}
    \mbox{AP}_K(R_q,G_q),
\end{equation}
where $Q$ is the seed entity sets for each query $q$. Given the ground-truth list $G_q$, $\mbox{AP}_K(R_q,G_q)$ denotes the average precision at position $K$ with the ranked list $R_q$.

\noindent\textbf{4. Implementation Details.}
For \modelName{}, we use OPT-2.7B~\cite{zhang2022opt} as our pre-trained language model. All the codes are implemented by Pytorch~\cite{paszke2019pytorch} and Huggingface’s Transformers~\cite{wolf-etal-2020-transformers} library. When utilizing the language model to generate entities, we choose the beam size $b = 30$ in our experiments. As for the Ranking step, we set the ranking weight $\lambda = 0.9$. Additionally, we provide a parameter study to explore the effect of these hyper-parameters on the expansion performance.
All experiments in this paper are conducted on a Linux server with Intel Xeon 6326 CPU and 256GB of RAM. Each experiment is assigned one Nvidia RTX 3090 GPU with 24GB of video memory to support model training and inference.

\subsection{Experiment Results}

\noindent\textbf{1. Overall Experimental Results.}


The overall performance of various ESE methods is shown in Table~\ref{tab:allresulta}.
Experimental results indicate that \modelName{} outperforms current state-of-the-art methods and other baselines on four datasets, demonstrating the effectiveness of \modelName{}.
Additionally, Table~\ref{tab:speed} shows the total elapsed time of different methods for expansion.
The time consumption of \modelName{} is considerably less than that of other methods, showing its outstanding efficiency.

\begin{table}[ht]
\centering
\caption{Expansion time of different methods. we select several strong baselines with publicly available codes for comparison. Each experiment is run 3 times and the average duration is computed as the final result.}
\scalebox{0.85}{
\begin{tabular}{lcccc}
\toprule
\textbf{Methods} & \textbf{Wiki} & \textbf{APR} & \textbf{CoNLL} & \textbf{OntoNotes} \\ \midrule
SetExpan & 6600s & 5247s & 231s & 605s \\
CGExpan & 3884s & 10591s & 178s & 1013s \\
ProbExpan & 149156s & 428877s  & 13250s & 53774s\\
BootstrapGAN & 37174s & 27632s & 1023s & 2997s\\ \midrule
\modelName{} & 2219s & 907s & 128s & 452s \\
\modelName{} (per Query) & 55.5s & 61.1s & 32.0s & 41.1s \\
\bottomrule
\end{tabular}
}
\label{tab:speed}
\end{table}

\noindent\textbf{2. Effectiveness and Efficiency Analysis.}
(1) For MAP@$K$ metrics on various datasets, \modelName{} consistently performs at a competitive level. 
On the relatively simple Wiki, APR datasets, ProbExpan employs self-supervised contrastive learning to optimize entity representation and outperforms other baselines. Our method fully utilizes the potential of pre-trained autoregressive language model to identify semantics of entities, also resulting in remarkable performance, achieving over 99\% on MAP@10 and MAP@20, even surpassing ProbExpan.

\begin{table}[ht]
    \centering
        \small
        \caption{Ablation results on Wiki and APR. We report the MAP@K(K=10/20/50) metrics.\modelName{} is composed of OPT + Cl(ass Name) + Ca(libration) + Ra(nking). The Ranking part should be constructed based on Class name part.}
    \scalebox{0.85}{
    \begin{tabular}{lcccccc}
    \toprule \multirow{2}{*} { \textbf{Methods} } & \multicolumn{3}{c} { \textbf{Wiki} } & \multicolumn{3}{c} { \textbf{APR} }\\ 
    \cmidrule(r){2-4} \cmidrule(r){5-7}
    & K = 10 & K = 20 & K = 50 & K = 10 & K = 20 & K = 50 \\
    \midrule 
    OPT & 91.2 & 86.7 & 78.5 & 86.6 & 81.8 & 72.9 \\
     + Ca & 93.1 & 89.2 & 82.9 & 88.3 & 84.8 & 80.4 \\
     + Cl & 94.4 & 91.5 & 85.7 & 92.6 & 88.3 & 83.7 \\
     + Cl + Ca  & 96.2 & 93.8 & 87.1 & 94.1 & 91.2 & 87.6 \\
     + Cl + Ra  & 98.5 & 97.4 & 90.8 & 97.5 & 96.8 & 92.4 \\
     \modelName{}   & \textbf{99.8} & \textbf{99.0} & \textbf{93.5} & \textbf{100.0} & \textbf{99.8} & \textbf{96.9}\\
    \midrule
    \end{tabular}
    }
\label{tab:ablation}
\end{table}

On the more challenging OntoNotes dataset, where differences among various semantic classes are relatively small, BootstrapGAN which models the bootstrapping process and the boundary learning process in a GAN framework is more effective. 
Since Class Name Generation strengthens the supervision signal and Generative Ranking adds discriminative conditions, our method is skilled at identifying boundaries among semantic classes and accessing knowledge within pre-trained model, even for challenging-to-distinguish entities.
Consequently, our method also achieves impressive performance, even outperforming BootstrapGAN's results.

(2) For time efficiency on different datasets, our approach shows a substantial advantage over other baselines. Compared to CGExpan, \modelName{} gains 75\% and 1068\% speedup on Wiki and APR respectively, while compared to BootstrapGAN, \modelName{} gains 699\% and 563\% speedup on CoNLL and OntoNotes. The efficiency advantage is even more pronounced compared to the current SOTA ProbExpan.
Note that CGExpan and ProbExpan consume more time on APR than on Wiki, while our approach exhibits the opposite trend. The reason is that APR contains more candidate entities and larger corpus, while Wiki contains more queries, and our generative approach, unlike previous retrieval-based approaches, is not affected by the number of candidate entities and the size of corpus.

We calculate the average expansion time per query for \modelName{}, finding it to be similar for both Wiki and APR, as well as for CoNLL and OntoNotes. The difference in average expansion time between these groups (i.e., Wiki and APR, CoNLL and OntoNotes) is attributed to fewer expansion iterations on CoNLL and OntoNotes.

(3) The previous strong baseline models (e.g., BootstrapGAN, ProbExpan) lack generalization ability. BootstrapGAN assumes that each candidate entity belongs to a fixed set of classes, which holds true in CoNLL and OntoNotes datasets but not in Wiki, APR datasets and real-world scenarios. In Wiki and APR, a large number of candidate entities do not fall within any predefined classes. As a result, this assumption leads to poor performance.
Meanwhile, ProbExpan acquires entity representation through contrastive learning  based on the specific corpus to each dataset.
However, for CoNLL and OntoNotes, the corpus size is relatively small, and the differences among various semantic classes are also quite small.
In such situation, ProbExpan, which heavily relies on corpus, encounters significant difficulties in learning semantic features of entities.

\begin{figure}[ht]
\centering
\setlength{\abovecaptionskip}{0.5em}
\subfloat[Wiki-Model Size] 
{ \label{wikim} 
\includegraphics[height = 0.39 \columnwidth, width=0.46\columnwidth]{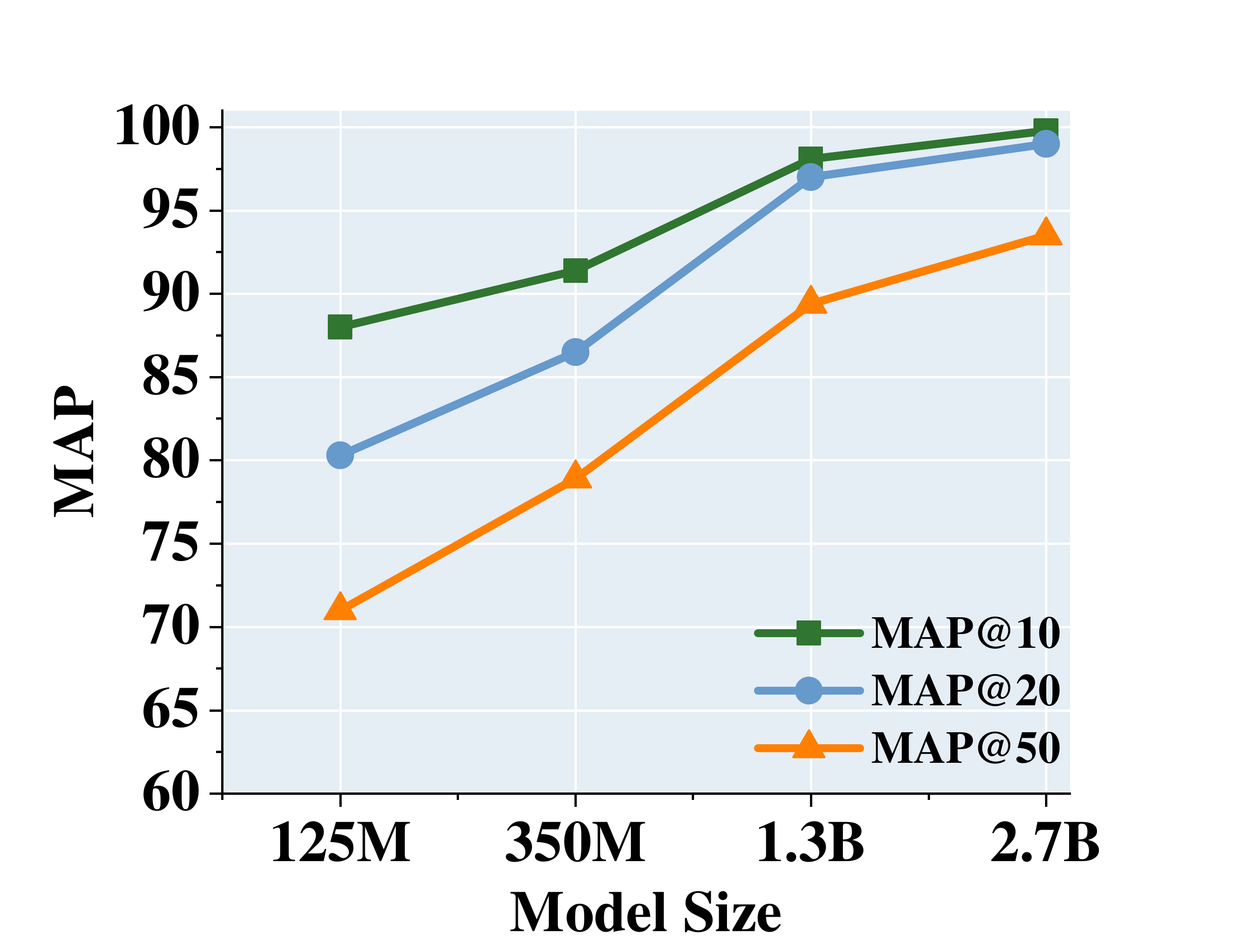} 
} 
\subfloat[APR--Model Size] 
{ \label{aprm} 
\includegraphics[height = 0.39 \columnwidth,width=0.46\columnwidth]{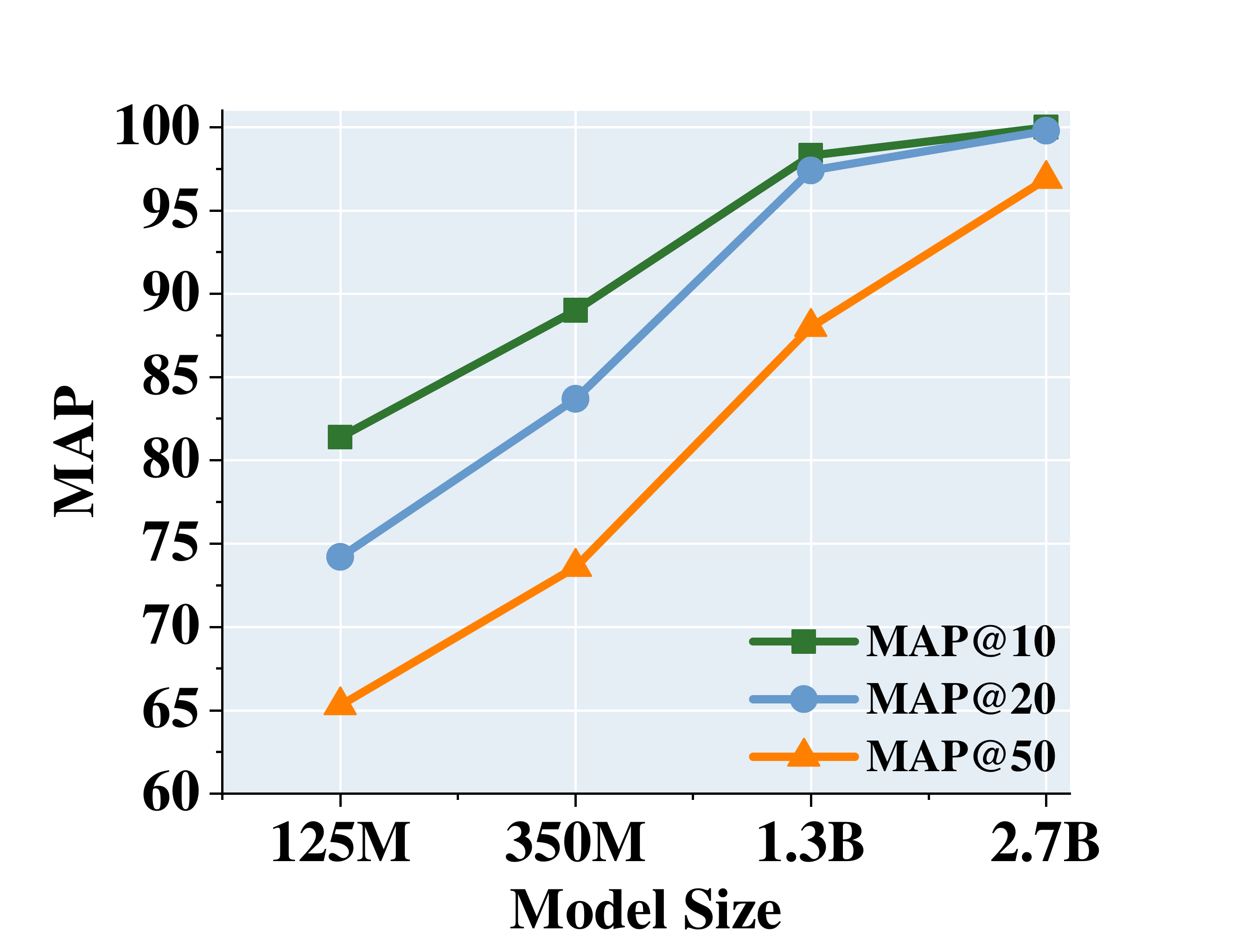} 
} 

\subfloat[Wiki-Beam Size] 
{ \label{wikib} 
\includegraphics[height = 0.39 \columnwidth, width=0.46\columnwidth]{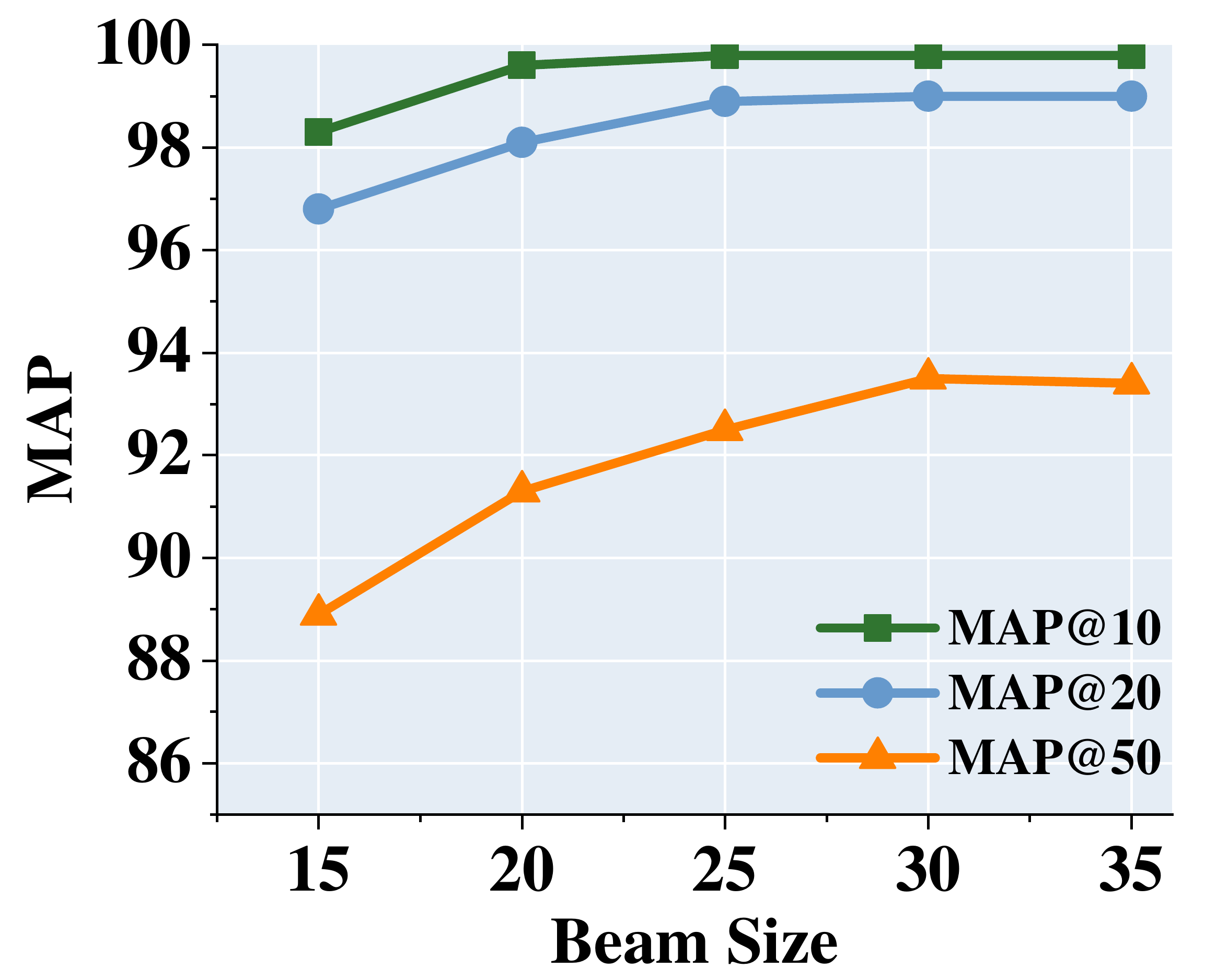} 
} 
\subfloat[APR-Beam Size] 
{ \label{aprb} 
\includegraphics[height = 0.39 \columnwidth, width=0.46\columnwidth]{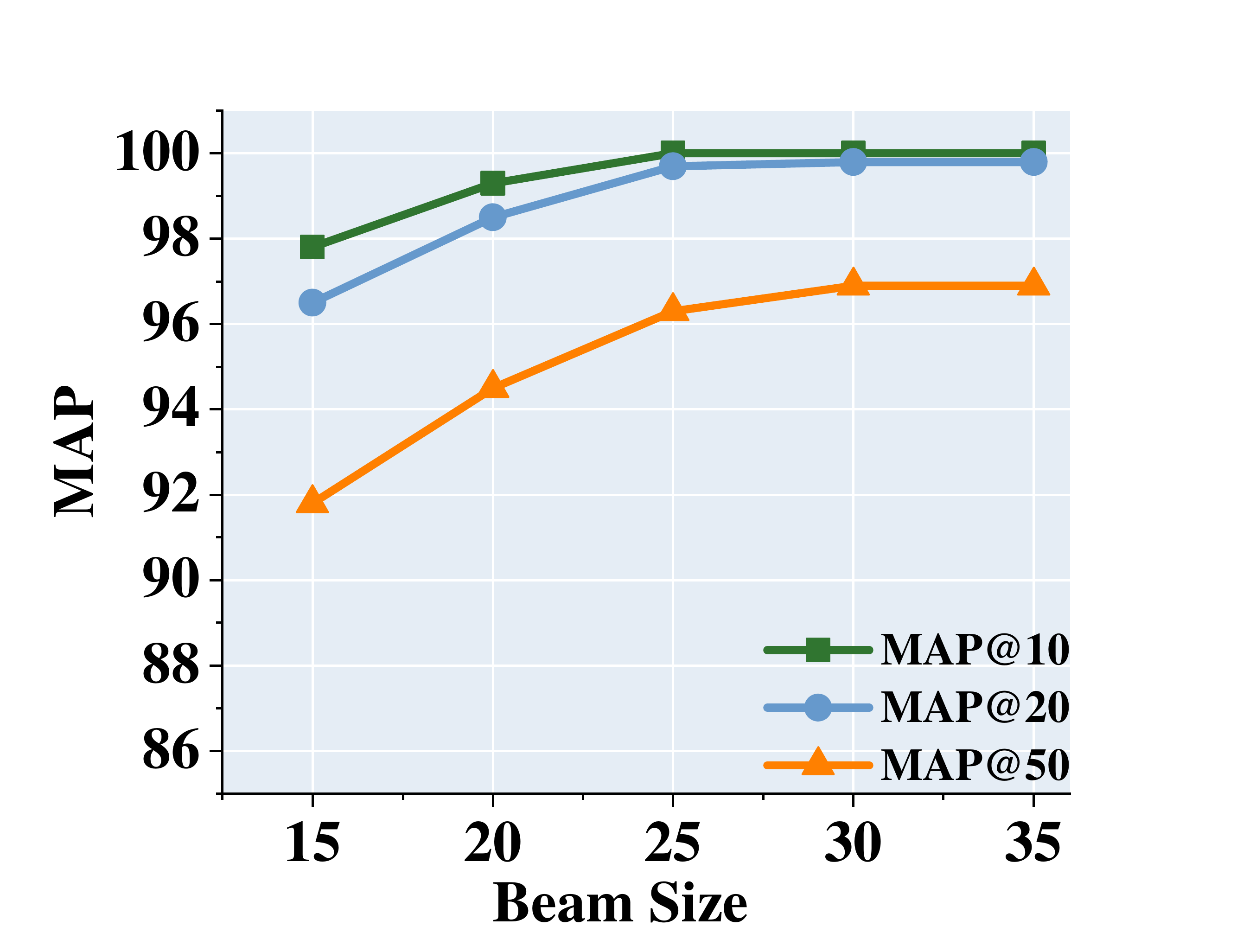} 
}

\subfloat[Wiki-Ranking weight] 
{ \label{wikip} 
\includegraphics[height = 0.39 \columnwidth, width=0.46\columnwidth]{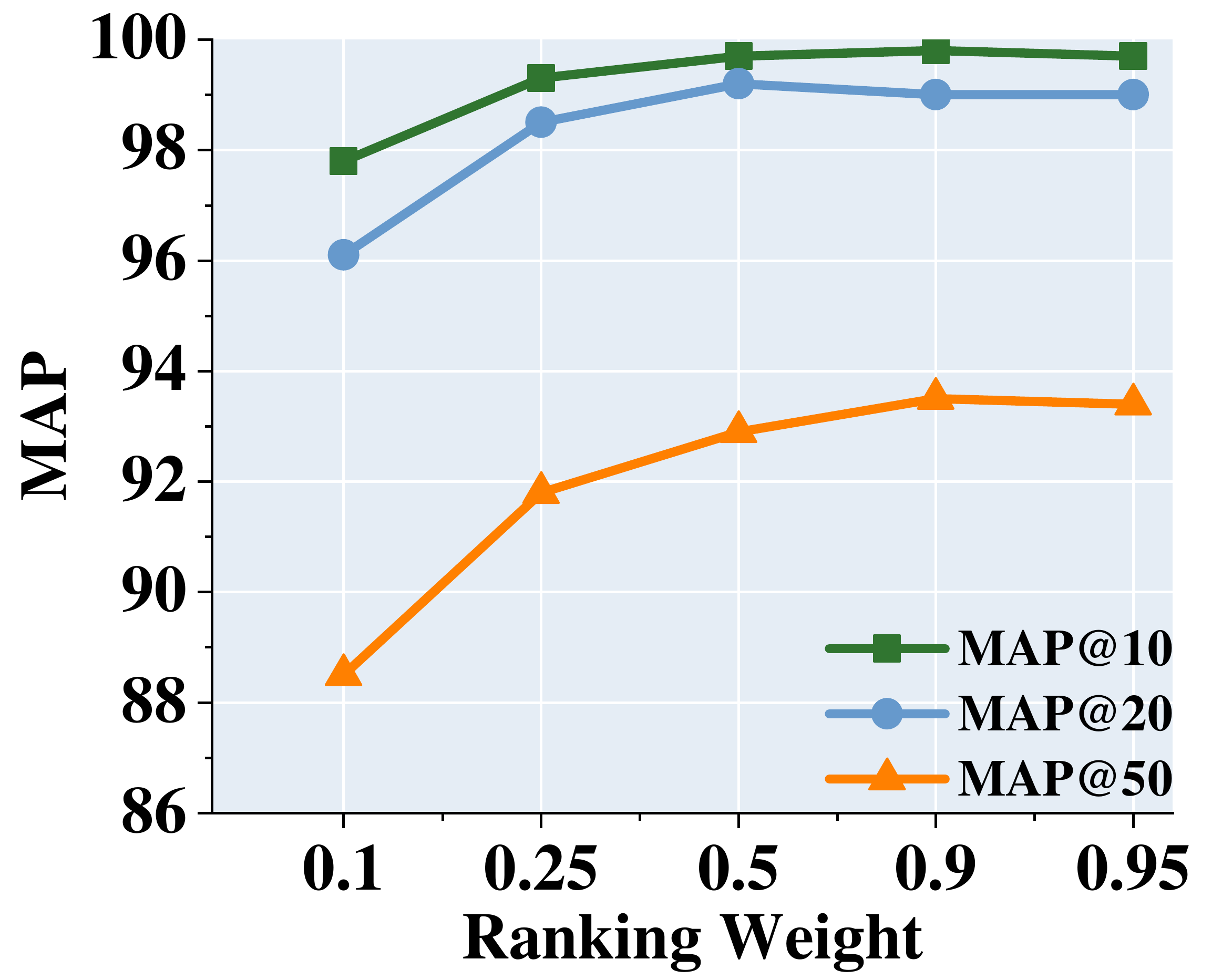} 
}
\subfloat[APR-Ranking weight] 
{ \label{aprp} 
\includegraphics[height = 0.39 \columnwidth, width=0.46\columnwidth]{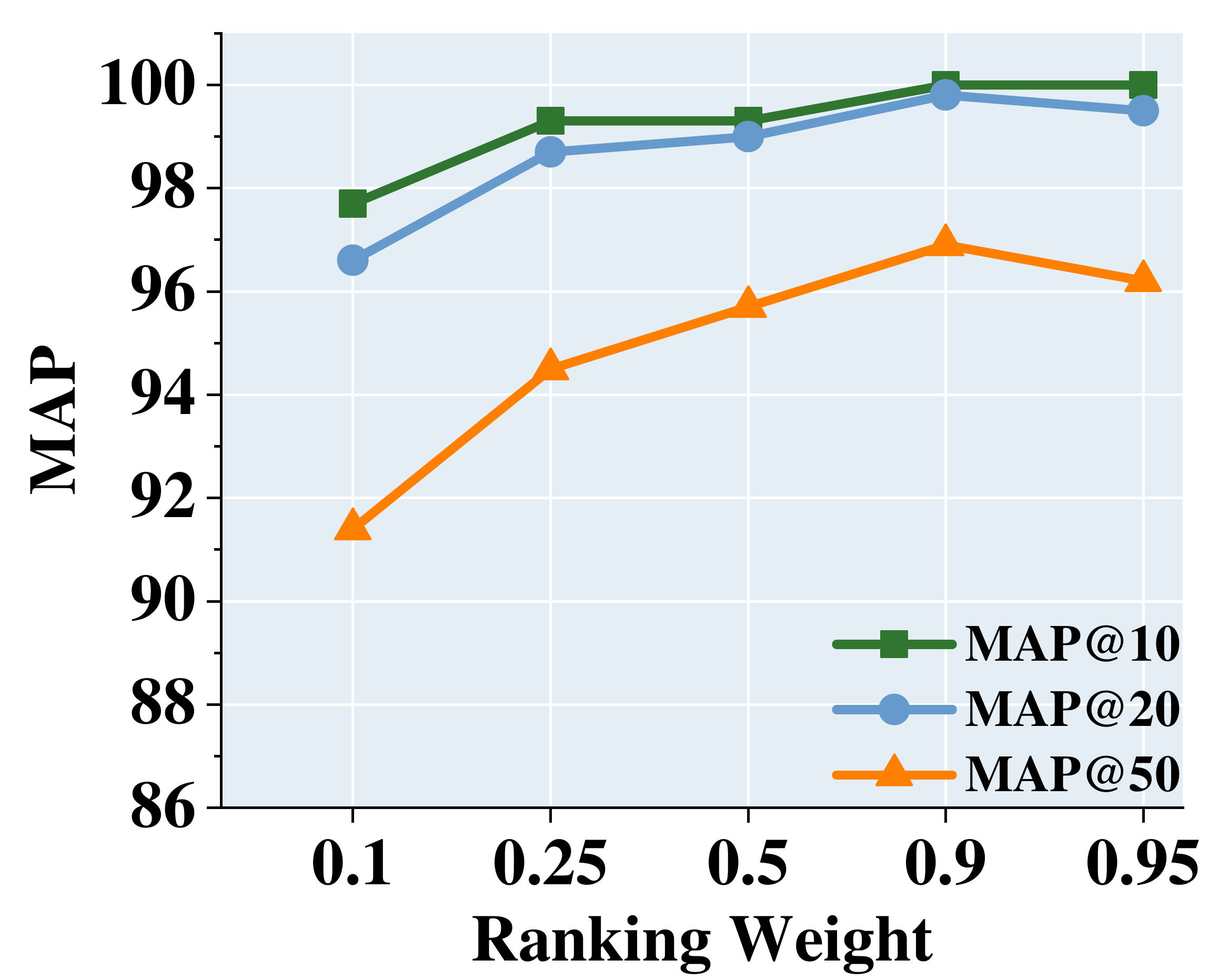} 
}
\caption{Parameter sensitivity analysis of model size, beam size and ranking weight in \modelName{}.} 
\label{tab:sensitive} 
\end{figure} 

On the contrary, our corpus-independent generative ESE approach exhibits high generalization ability. 
Although these four datasets provide different corpora, our approach easily adapts to different datasets.
Moreover, our approach can also be generalized to the scenarios where corpus availability is limited, making it suitable for a broader range of downstream tasks.

\subsection{Ablation Studies}

\begin{figure*}[ht]
\centering
\includegraphics[width=0.85\textwidth]{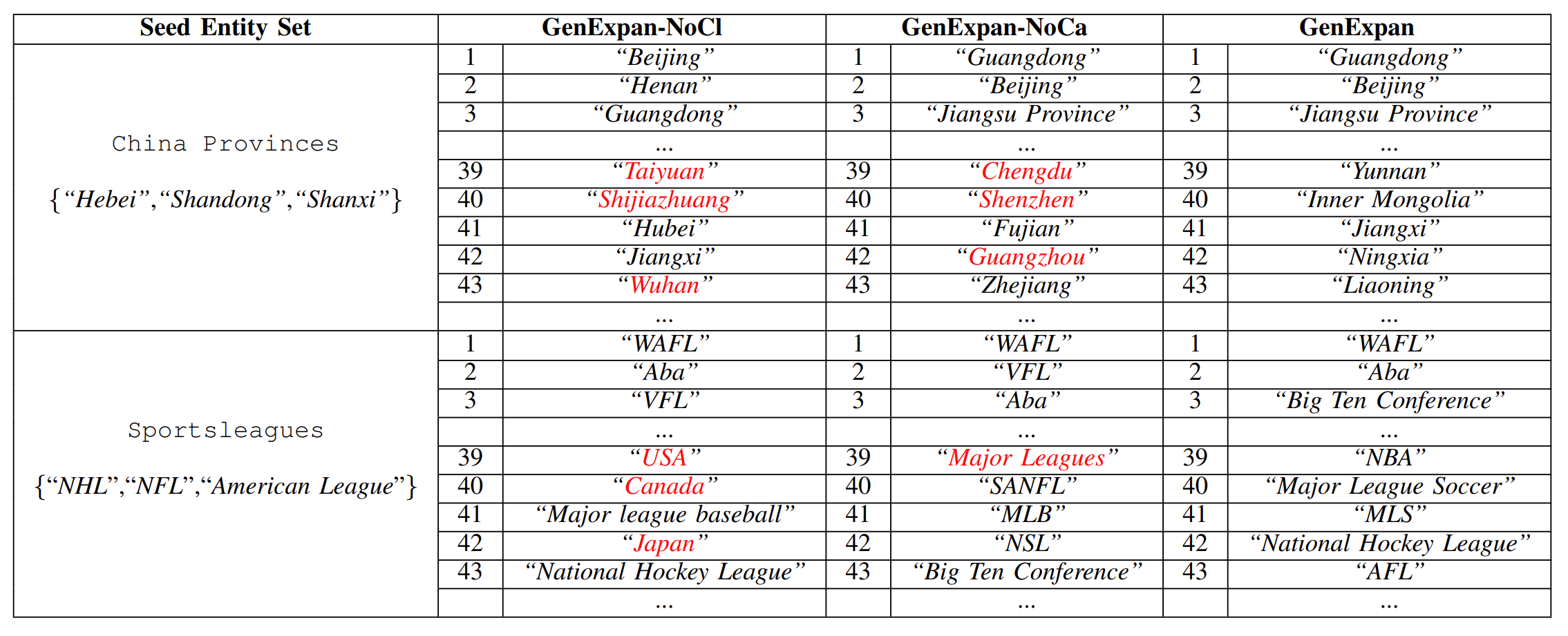}
\caption{We select \texttt{China Provinces} and \texttt{Sportsleagues} classes because they lead to poor performance in case of no Ca(libration) and no Cl(ass Name), as shown in Table~\ref{tab:ablation}. We mark the wrong entities in red.}
\label{tab:calibration p}
\end{figure*}

\begin{table*}[ht]
\small
\centering
\caption{Class name generated for corresponding seed entities.}
\scalebox{0.9}{
\begin{tabular}{|c|c|c|}
\hline
\textbf{Seed Entities}  & \textbf{Generated Class Name} & \textbf{Ground Truth Class Name} \\ \hline
\makecell[c]{\emph{``ABC Sports'', ``ESPNews'', ``The Learning Channel''}} & Cable TV channels & TV channels \\ \hline
\makecell[c]{\emph{``Hebei'', ``Shandong'', ``Shanxi''}} & Provinces in China & Chinese provinces\\  \hline
\makecell[c]{\emph{``Republican'', ``Liberals'', ``Conservatives''}} & Political parties & Parties \\  \hline
\makecell[c]{\emph{``United States'', ``Canada'', ``China''}} & Countries & Countries\\  \hline
\makecell[c]{\emph{``NHL'', ``NFL'', ``American League''}} & Professional sports leagues & Sports leagues \\  \hline
\makecell[c]{\emph{``Illinois'', ``Arizona'', ``California''}} & US states & US states\\  \hline
\makecell[c]{\emph{``Google'', ``Goldman Sachs'', ``Facebook''}} & Companies & Companies\\  \hline
\makecell[c]{\emph{``Heart attacks'', ``AIDS'', ``Strokes''}} & Diseases & Diseases \\  \hline
\end{tabular}
}
    \label{tab:cln generation}
\end{table*}

To provide a comprehensive analysis of \modelName{}'s effectiveness, we conduct a series of ablation experiments to investigate how each module impacts the expansion performance. The results and meanings of \modelName{}'s variants are presented in Table~\ref{tab:ablation}.

\noindent\textbf{1. Can Class Name Enhance Semantic Supervision?}

As shown in Table~\ref{tab:ablation}, OPT + Class Name outperforms OPT, particularly for challenging MAP@20/50 metrics, indicating that Class Name is highly beneficial to the overall performance.
In experiments, we observe that if class name information is not included, the model often generates entities that are associated with seed entities but do not belong to the same semantic class.
Therefore, incorporating the class name information indeed enhances semantic supervision, and ensures generated entities meet requirements of ESE task.
In Case Studies (Section~\ref{sec:case}), we provide further evidence.

\noindent\textbf{2. Can Knowledge Calibration Work?}


We posit that Knowledge Calibration alleviates the over-preference of pre-trained generation model towards common entities, thereby reducing the gap between the model's generic knowledge and the requirements of the ESE task.
In Table~\ref{tab:ablation}, we provide a clear and detailed comparison between variants with and without Calibration. This analysis offers compelling evidence of Calibration's remarkable impact.
Moreover, we use practical examples to offer a comprehensive analysis of this phenomenon in Case Studies (Section~\ref{sec:case}).

\noindent\textbf{3. Can Generative Ranking Empower ESE?}


Generative Ranking calculates the probability of generating seed entities and class names, considering detailed ranking information from the model's perspective.
In Table~\ref{tab:ablation}, the comparison between \modelName{} and OPT + Class Name + Calibration clearly illustrates that Generative Ranking plays a significant role in refining entity expansion. Additionally, the superior performance of OPT + Class Name + Ranking over OPT + Class Name also underscores the effectiveness of Generative Ranking in empowering ESE task.

\subsection{Parameter Studies}
We conduct a thorough exploration of three key parameters within \modelName{}, including the model size, the beam size $b$ and the weight of log probability $\lambda$ in generative ranking. We analyze how these hyper-parameters affect the overall performance in detail.

\noindent\textbf{1. Model Size.} 
The core of \modelName{} is the expansion-oriented text generation process. 
A larger language model has more substantial text generation capability.
We explore the effect of model size on the experiment results.
Figure~\ref{wikim} and~\ref{aprm} illustrate that as the model size increases, the performance of the model rises noticeably. 
Due to the limitation of model capacity, small models (e.g. 125M, 350M) cannot generate accurate class names based on In-context Learning, and they can hardly generate similar uncommon entities from seed entities, resulting in poor performance.
Limited by hardware resources, we cannot explore models with larger size.

\noindent\textbf{2. Beam Size.} 
We also explore the impact of beam size in entity generation on performance. From Figure~\ref{wikib} and~\ref{aprb}, we observe that an increase in beam size leads to increased expansion performance. 
In the decoding process of Beam Search, a smaller beam size implies a smaller search space. When using a small beam size, the model faces challenges in generating diverse entities and long entities (more than 3 tokens).
Figure~\ref{wikib} and~\ref{aprb} also show that the model performance changes slightly when beam size $b\ge 25$, which indicates that the performance of our approach is not sensitive to this hyper-parameter when it is within a reasonable range.

\noindent\textbf{3. Ranking Weight.} 
From Figure~\ref{wikip} and Figure~\ref{aprp}, we observe that the expansion performance promotes as the ranking weight increases.
The results show that Generative Ranking can better reveal the semantic similarity between generated entities and seed entities rather than only using entities occurrences.
After the ranking weight is increased to a certain amount, the performance change is minimal, reflecting that the expansion performance is not sensitive to the ranking weight within a reasonable range (from 0.5 to 0.95).

\subsection{Case Studies}\label{sec:case}

\begin{figure*}[ht]
\centering
\includegraphics[width=0.8\textwidth]{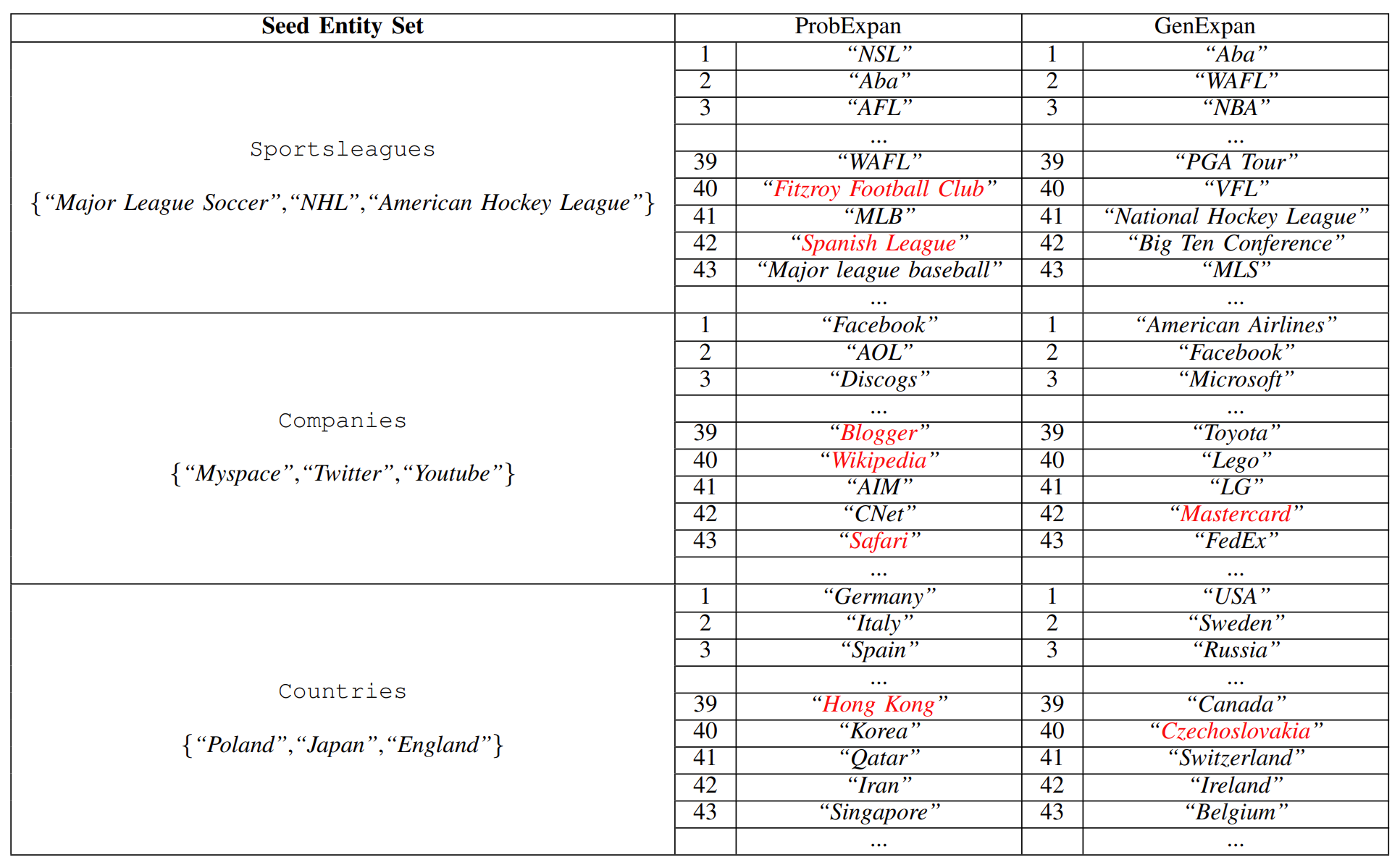}
\caption{Comparison between ProbExpan and \modelName{}. We mark wrong entities in red.}
\label{tab:cmp to baseline}
\end{figure*}

\noindent\textbf{Class Name Generation.}
GenExpan-NoCl and GenExpan columns of 
Figure~\ref{tab:calibration p} reveal that incorporating class names improves the supervision of entity category.
With the introduction of ``Professional sports leagues'' as the class name, the likelihood of erroneously expanding certain country names (e.g., ``USA'', ``Canada'') based on the seed entity ``American League'' is significantly reduced.
Moreover, Table~\ref{tab:cln generation} shows the generated class names for several queries from different semantic classes.
We observe that most generated class names have the same meanings as the ground truth class names. 

\noindent\textbf{Knowledge Calibration.}
From the GenExpan-NoCa column of 
Figure~\ref{tab:calibration p}, we can see the language model's tendency to generate common but incorrect words. For instance, both ``Shenzhen'' and ``Guangzhou'' are common Chinese cities but not Chinese provinces. Therefore, they are more likely to be expanded rather than unusual but correct entities (e.g., ``Inner Mongolia'', ``Jiangxi'') because of the language model bias. The calibration alleviates the model's bias, resulting in correct entities that are in line with semantic classes. 

\noindent\textbf{Compared to Strongest Baseline.}
Based on the results presented in 
Figure~\ref{tab:cmp to baseline}, model mitigates expansion errors resulting from reliance on common sense.
For instance, a person lacking external background knowledge may erroneously classify ``Fitzroy Football Club'' and the ``Spanish League'' into the ``Sportsleagues'' category.
The semantic representations of rare entities may be inaccurate in ProbExpan.
This leads to errors, similar to how human misinterprets words solely based on their surface form.
The phenomenon highlights the advantages of our corpus-independent framework, as \modelName{} fully leverages the pre-trained model's vast knowledge.

\noindent\textbf{Error Cases.}
Interestingly, we find that \modelName{} sometimes incorrectly expands some entities (e.g., \emph{``Czechoslovakia''} and \emph{``MasterCard''}) as shown in Figure~\ref{tab:cmp to baseline}. 
``Czechoslovakia'' is a country that no longer exists today, splited in 1993. We think that the model we use expands it as a country may be related to its own pre-training process. During the model's pre-training, it encounters some previous linguistic data that treats ``Czechoslovakia'' as a country.
Furthermore, in Wiki's corpus, ``MasterCard'' is consistently referred to a credit card. Therefore, Probexpan, which depends on the semantics within the corpus, doesn't regard ``MasterCard'' as a company. Our approach, being independent of the corpus, tends to view ``MasterCard'' as a company, which is also a valid perspective.

\section{Conclusions}
Most existing retrieval-based ESE methods iteratively traverse the corpus and the entity vocabulary, leading to poor efficiency and difficulty in applying on large-scale data scenarios. 
In this paper, we propose a novel corpus-independent Generative Entity Set Expansion framework, \modelName{}.
\modelName{} consists of four parts, Class Name Generation, Prefix-constrained Entity Generation, Knowledge Calibration and Generative Ranking, which utilizes text generation ability of pre-trained auto-regressive language model to accomplish ESE task.
The experiments indicate that \modelName{} outperforms competitors across four datasets both in expansion performance and time efficiency.
In the future, we will further study the factors of prompt and explore the impact of these factors. 
Moreover, it is also a promising research direction to utilize generative expansion under more actual but challenging scenarios, such as non-named entities, multifaceted entities, and vague concepts.

\section*{Acknowledgments}
This research is supported by National Natural Science Foundation of China (Grant No. 62276154), Research Center for Computer Network (Shenzhen) Ministry of Education, the Natural Science Foundation of Guangdong Province (Grant No. 2023A1515012914), Basic Research Fund of Shenzhen City (Grant No. JCYJ20210324120012033 and GJHZ202402183000101), the Major Key Project of PCL for Experiments and Applications (PCL2021A06).

\bibliographystyle{ACM-Reference-Format}
\bibliography{sample-base}










\end{document}